\documentclass[11pt,a4paper]{article}

\usepackage[hyperref]{emnlp2020}
\usepackage{times}
\usepackage{latexsym}

\usepackage{booktabs}
\usepackage{subfigure}
\usepackage{graphicx}
\usepackage{amsmath,amsfonts,amssymb,amsthm,color}
\usepackage{cleveref}
\usepackage[normalem]{ulem}

\usepackage{paralist}
\usepackage{tablefootnote}

\usepackage{microtype}

\usepackage[mathscr]{eucal}
\usepackage[inline]{enumitem}
\usepackage{algorithm,algorithmic}

\newcommand{\BCal}{\mathscr{B}}
\newcommand{\NCal}{\mathscr{N}}
\newcommand{\RCal}{\mathscr{R}}
\newcommand{\SCal}{\mathscr{S}}
\newcommand{\XCal}{\mathscr{X}}
\newcommand{\YCal}{\mathscr{Y}}

\newcommand{\xvec}{\mathbf{x}}
\newcommand{\yvec}{\mathbf{y}}

\aclfinalcopy % Uncomment this line for the final submission
\title{Semantic Label Smoothing for Sequence to Sequence Problems}

\author{
    Michal Lukasik, 
    Himanshu Jain, 
    Aditya Krishna Menon, 
    Seungyeon Kim,\\
    {\bf Srinadh Bhojanapalli, 
    Felix Yu,
    Sanjiv Kumar}\\
  Google Research \\
  \texttt{mlukasik@google.com}}

\date{}

\begin{document}
\maketitle

\begin{abstract}
Label smoothing has been shown to be an effective regularization strategy in classification, that prevents overfitting and helps in label de-noising.
However, extending such methods directly to seq2seq settings, such as Machine Translation, is challenging: 
the large target output space of such problems makes it intractable to apply label smoothing over all possible outputs. 
Most existing approaches for seq2seq settings either do token level smoothing, or smooth over sequences generated by randomly substituting tokens in the target sequence. Unlike these works, in this paper, we propose a technique that smooths over \emph{well formed} relevant sequences that not only have sufficient n-gram overlap with the target sequence, but are also \emph{semantically similar}. Our method shows a consistent and significant improvement over the state-of-the-art techniques on different datasets.

\end{abstract}

\section{Introduction}
Label smoothing is a regularization technique commonly used in deep learning ~\citep{Szegedy:2016,Chorowski:2017,Vaswani:2017,Zoph:2018,Real:2018,Huang:2019}, that improves calibration~\citep{Muller:2019} and helps in label de-noising \citep{lukasik2020does}.
Here, one \emph{smooths} labels by introducing a prior in the label space (often just a uniform distribution) in order 
to prevent overly confident predictions
and achieve better model calibration, 
both of which lead to better generalization.

Given these benefits, it is natural to consider whether label smoothing can be applied to \emph{sequence-to-sequence} (seq2seq) prediction tasks in Natural Language Processing.
Here,
inducing a label prior involves smoothing in \emph{sequence space}. However, this is a challenging task because the output space is exponentially large for sequences, unlike the label space in standard classification. Previous works approached this challenge either by smoothing over individual tokens of the target sequence, or by sampling a few nearby targets according to Hamming distance or BLEU score \cite{NIPS2016_rmle,elbayad-etal-2018-token}. These techniques however do \emph{not} guarantee that the smoothed targets lie within the space of acceptable targets (i.e., the sampled new target may no longer be grammatically correct or even preserve semantic meaning).

In this work, we propose a label smoothing approach for seq2seq problems that overcomes this limitation.
Given a large-scale corpus of valid sequences,
our approach selects a subset of sequences 
that are not only semantically similar to the target sequence,
but also \emph{well formed}. 
We achieve this using a pre-trained model to find semantically similar sequences from the corpus, and then use BLEU scores to rerank the closest targets. 
We empirically show that this approach improves over competitive baselines on multiple machine translation tasks.

\section{Related Works}

% {\bf\color{olive} Sentence vs sequence}

\paragraph{Token-level smoothing} A popular approach used in language tasks is so called token level smoothing, where for each position's classification loss, a prior distribution over the entire vocabulary (uniformly or with unigram probability estimates) is used for regularization \citep{Pereyra:2017, edunov2017classical}. 
This is similar to the classical label smoothing (e.g.~\cite{Szegedy:2016}), as it smooths each token label independent of their context and position in the sequence.
Such an approach is thus unlikely to result in semantically related targets.

\paragraph{Sequence-level smoothing}

\citet{NIPS2016_rmle} augment the loss with a term rewarding predictions of sampled sequences. The sampling of sequences is based on their edit distance or Hamming distance to the target. This method thus smooths the loss over to similar sequences (in terms of the edit distance) with smoothing rewards. \citet{elbayad-etal-2018-token} employ a similar technique, but with a new reward function based on BLEU \cite{papineni2002bleu} or CIDEr \cite{vedantam2015cider} score. 
Specifically,
\citet{elbayad-etal-2018-token} %explored seq2seq label smoothing by
generate a smoothed version of the target sequence,
wherein one replaces a token with a random token (with up-sampling of rare words). 
Such newly generated sequences were given a partial reward based on the cosine similarity between the two tokens in a pretrained word-embedding space. 
This differs from our approach because this context-independent perturbation is limited to generating the same structure for the new sequence as that of the original sequence. 

%\seungyeonk{
\citet{Zheng:2018}, on the other hand, constructed grammatically correct and meaning preserving sequences.
%} \himj{
However, unlike our work, their approach relies on having multiple references (target sequences per input sequence) and might not be able to generate sequences where common words or synonyms do not appear in the same order, which is a strong limitation, precluding an augmentation like: \emph{Yesterday, he scored a 94 on his final} (original sequence), \emph{He had 94 points in the final test yesterday} (augmented sequence).
%}

% \himj{Other approaches such as \citet{Zheng:2018} do ensure grammatical consistency but it relies on having multiple references (target sequences per input sequence) and might not be able to generate sequences where common words/synonyms don’t appear in the same order, for instance: \emph{Yesterday, he scored a 94 on his final}, \emph{He had 94 points in the final test yesterday}.}

More broadly, an important shortcoming of such approaches is that sequences deemed close can actually lack important properties such as preserving the meaning of the original sequence. In particular, swapping even a single token in a sequence may cause a drastic shift in its meaning (e.g., turning a factually correct text into a false one) even though being close in the Hamming space. We address this shortcoming by restricting augmented target sequences to the training set, and selecting sequences based on similarity obtained from a pretrained model.

%\seungyeonk{
Unlike other approaches, \citep{bengio2015scheduled} proposed a scheduled sampling technique that does not depend on any external data source. Instead, it utilizes the self-generated sequences from the current model. Both our approach and the scheduled sampling technique bear similarity in that they aim at improving model generalization, by either providing semantically similar candidates (ours) or self-generated sequences (theirs). Indeed, these two approaches could complement each other by providing various ways of related but not exact targets.
%}

\paragraph{Hard negative mining}
Our work is also related to hard negative mining approaches that select a subset of confusing negatives for each input \cite{mikolov2013distributed, reddi2019stochastic, guo-etal-2018-effective}. Different from the above, we add a soft objective function over the sampled (relevant) target sequences, rather than treating them as negatives in the classification sense.

\section{Method}

Sequence-to-sequence (seq2seq) learning involves learning a mapping  
from an input sequence $\xvec$ (e.g., a sentence in English)
to an output sequence $\yvec$ (e.g., a sentence in French).
Canonical applications include machine translation
and question answering.
%There has been significant progress in this area in the last decade~\citep{Sutskever:2014}.

Formally,
let $\XCal$ denote the space of input sequences (e.g., all possible English sentences),
and
$\YCal$ the space of output sequences (e.g., all possible French sentences).
We represent by
$\xvec = [x_1, x_2, ... x_N]$
an input sequence consisting of $N$ tokens,
and 
similarly
$\yvec = [y_1, y_2, ... y_{N'}]$
an output sequence with $N'$ tokens.
Our goal is to learn a function $f \colon \XCal \to \YCal$
that, given an input sequence,
generates a suitable target sequence.

To achieve this goal, we have a training set 
%$\SCal = \{ ( \xvec_n, \yvec_n ) \}$
$\SCal \subseteq ( \XCal \times \YCal )^{n}$
comprising pairs of input and output sequences.
We then seek to minimise the objective
%The seq2seq task can be often defined as the following:
\begin{equation}
    \label{eqn:base-loss}
    %\min_{f} \sum_{( \xvec, \yvec ) \in \SCal} \ell(\xvec, \yvec) = l(f(\xvec), \yvec),
    L( \theta ) = \sum_{( \xvec, \yvec ) \in \SCal} -\log p_{\theta}( \yvec \mid \xvec ),
\end{equation}
where
$p_{\theta}( \cdot | \xvec; \theta )$ is a parametrized distribution over all possible output sequences.
% and
% $\ell( \hat{\yvec}, \yvec )$ is a \emph{loss function}.
Given such a distribution,
we choose $f( \xvec ) = \operatorname{argmax}_{\yvec \in \YCal} p_{\theta}( \yvec \mid \xvec ) $.
Observe that one may implement~\eqref{eqn:base-loss} via a token-level decomposition,
$$ L( \theta ) = \sum_{( \xvec, \yvec ) \in \SCal} \sum_{i = 1}^N -\log p_{\theta}( y_i \mid \xvec, y_1, \ldots, y_{i - 1} ). $$
This may be understood as a maximum likelihood objective,
or equivalently
the \emph{cross-entropy} between 
$p_{\theta}( \cdot | \xvec; \theta )$ and a one-hot distribution concentrated on $\yvec$.

% Often, this is the cross-entropy function over the entire sequence:
% \begin{align*}
%     \ell(\hat{\yvec}, \yvec) = \sum_{i=1}^N \sum_{v \in \mathcal{V}} ( \mathbf{1}_{y_i = v} \log [\hat{\yvec}_{i} ]_v ),
% \end{align*}
% where $\mathcal{V}$ is the vocabulary.

%
\noindent \textbf{Label smoothing meets seq2seq.}
Intuitively, the cross-entropy objective encourages the model to score the \emph{observed} sequence $\yvec$ higher than any ``competing'' sequence $\yvec' \neq \yvec$.
While this is a sensible goal,
one limitation observed from classification settings
is that the loss may lead to models that are overly confident in their predictions,
which can hamper generalisation~\citep{Guo:2017}.

\emph{Label smoothing}~\citep{Szegedy:2016,Pereyra:2017,Muller:2019} is a simple means of correcting this in classification settings.
Smoothing involves simply adding a small reward to all possible \emph{incorrect} labels, i.e.,
mixing the standard one-hot label with a uniform distribution over all labels.
This regularizes the training and generally leads to better predictive performance as well as probabilistic calibration~\citep{Muller:2019}.

% \label{sec:model_arch}
% Let us consider structured: input $\xvec$ and output $\yvec$.

%
Given the success of label smoothing in classification settings,
it is natural to explore its value in seq2seq problems.
However, standard label smoothing is clearly inadmissible:
it would require smoothing over all possible outputs $\yvec' \in \YCal$,
which is typically an intractably large set.
Nonetheless, we may follow the basic intuition of smoothing
by adding a subset of \emph{related} targets to the observed sequence $\yvec$, 
yielding a \emph{smoothed loss}
%We now extend this intuition for the seq2seq task by 
\begin{equation}
    %\numberthis
    \label{eqn:smoothing-seq2seq}
    \resizebox{0.875\linewidth}{!}{
    $\displaystyle
    -\log p_{\theta}( \yvec \mid \xvec ) + \frac{\alpha}{|\RCal( \yvec )|} \cdot \sum_{\yvec' \in \RCal( \yvec )} -\log p_{\theta}( \yvec' \mid \xvec ).
    $}
\end{equation}
Here, $\RCal( \yvec )$ is a set of \emph{related sequences} that are similar to the ground truth $\yvec$,
and $\alpha > 0$ is a tuning parameter that controls how much we rely on the observed versus related sequences.
%but not exactly the same.

The quality of $\RCal( \yvec )$ is important for our task.
Ideally, we would like an $\RCal( \yvec )$ that:
\begin{enumerate*}[label=(\roman*)]
    \item is efficient to compute, and
    \item comprises sequences 
    %bearing semantic similarity to $\yvec$.
    which meaningfully align with $\xvec$ (e.g., are plausible alternate translations).
\end{enumerate*}
We now assess several options for constructing  $\RCal( \yvec )$
in light of the above.

\noindent {\bf Random sequences.}
One simple choice is to choose a random subset of output sequences from the training set. 
In the common setting where $f$ is learned by minibatch SGD on randomly drawn minbatches $\BCal = \{ ( \xvec^{(i)}, \yvec^{(i)} ) \}$,
one may simply pick $\RCal( \yvec )$ to be all output sequences in $\BCal$.

Such random sequences contain general target language understanding (e.g., French grammar for an English to French translation task). 
However, these sequences are unlikely to have any semantic correlation with the true label.
% which limits the effectiveness of the approach.

%
\noindent {\bf Token-level smoothing.}
To ensure greater semantic correlation between the  selected sequences and the original $\yvec$,
one idea is to perform \emph{token-level smoothing}.
For example, 
~\citet{Vaswani:2017} proposed to smooth uniformly over all tokens from the vocabulary.
{\color{olive}~\citet{elbayad-etal-2018-token}} proposed to construct sequences $\yvec' = [ y'_1, y'_2, \ldots, y'_{N'} ]$
where
for a randomly selected subset of tokens $j \in [N']$,
$y'_i$ is some \emph{related} token in the minibatch;
for other tokens,
$y'_i = y_i$.
These related tokens are chosen so as to maximise the BLEU score between $\yvec$ and $\yvec'$.

While this approach increases the semantic similarity to $\yvec$, operating on a token level is limiting.
For example, one may change the meaning of a factual sentence by changing even a few words.
Further, operating at a per-token level limits the \emph{diversity} of $\RCal( \yvec )$,
since,
e.g.,
all sequences have the same number of tokens and structure as $\yvec$.

\vspace{2pt}
\noindent {\bf Proposal: semantic smoothing.}
To overcome the limitations of token-level smoothing,
we would ideally like to directly smooth over related \emph{sequences}.
Our basic idea is to seek sequences
$$ \RCal( \yvec ) = \{ \yvec' \colon s_{\mathrm{sem}}( \yvec, \yvec' ) \land s_{\mathrm{bleu}}( \yvec, \yvec' ) >  1 - \epsilon \}, $$
where $s_{\mathrm{sem}}$ is a score of semantic similarity,
and $s_{\mathrm{bleu}}$ is the BLEU score.
Intuitively, our relevant sequences comprise those that are both semantically similar to $\yvec$, \emph{and} have sufficient unigram overlap.
\renewcommand{\algorithmicrequire}{\textbf{Input:}}
\renewcommand{\algorithmicensure}{\textbf{Output:}}

\begin{algorithm}[!t]
    \caption{Sampling of related sequences.}
    \label{alg:proposal}
    \begin{algorithmic}[1]
        \REQUIRE example $( \xvec, \yvec )$; sequences $\YCal_{\mathrm{ref}}$
        \ENSURE related sequences $\RCal( \yvec )$
        
        \STATE Embed reference sequences, e.g., using BERT
        \STATE $\NCal( \yvec ) \leftarrow k$ closest sequences to $\yvec$ from $\YCal_{\mathrm{ref}}$ in the embedding space.
        \STATE Sort elements of $\NCal( \yvec )$ by BLEU score to $\yvec$.
        \STATE $\RCal( \yvec ) \leftarrow$ top $k'$ elements from $\NCal$.
    \end{algorithmic}
\end{algorithm}
\begin{table}[!t]
    \centering
    \scriptsize
    {
    \begin{tabular}{lr}
        \toprule
        Orig: & Yesterday, he scored a 94 on his final.\\
        \midrule
        1st: &  He had 94 points in the final test yesterday. \\
        2nd: &  But the child just scored 9 points on the Apgar test.\\
         \toprule
         Orig: & Exchange of experience and good practices.\\
         \midrule
         1st: & Exchange of best practices.\\
         2nd: & Exchange of information and best practices.\\
         \toprule
        Orig: & Nothing else I can do?\\
         \midrule
        1st: &Is there anything else I can do for you, sir?\\
        2nd: &Can I do something for you?\\
        \bottomrule
    \end{tabular}
    }
    
    \caption{English translations of top two augmentations from BERT+BLEU4 on examples from EN-CS.}
    \label{tbl:examples}
\end{table}
A key challenge is efficiently identifying semantically similar sequences to $\yvec$.
To achieve this in a tractable manner, we propose the following procedure (see Algorithm~\ref{alg:proposal}).
First, we assume the existence of an \emph{embedding space} for output sequences.
For example, this could be the result of BERT~\citep{Devlin:2019}, which embeds each sequence into a fixed vector representation.
Given such an embedding space
and a corpus $\YCal_{\mathrm{ref}}$ of reference sequences, 
we may now efficiently compute the neighbors of $\yvec$, $\NCal( \yvec )$, comprising the top-$k$ closest sequences in $\YCal_{\mathrm{ref}}$ for the given $\yvec$ \cite{ApproximateNearestNeighbors98}.\footnote{Alternatively, one could consider selecting highest scoring augmentations based on a pre-trained seq2seq model. However, the resulting quadratic computational complexity renders such an approach impractical.}
%of such approach is $O(N\timesM)$ of scorings, where $N$ is train data size and $M$ is the corpus size of the augmentation candidates,

\begin{table*}[!t]
    \centering
    {
    \begin{tabular}{llcccc@{}}
        \toprule
        Method & $\alpha$ &\textbf{EN-DE} & \textbf{EN-CS} & \textbf{EN-FR}\\
        \toprule
        Base setup  \cite{Vaswani:2017} & --- & 28.03 & 21.19  & 39.66 & \\
        Token LS \cite{Vaswani:2017,Szegedy:2016} & $0.1$ & 28.72 & 21.47 & 39.87\\ 
        Within batch sequence LS \cite{guo-etal-2018-effective} & $0.001$ & 28.81&  21.26 & 39.21\\
        Sampled augmentations BLEU4 \cite{elbayad-etal-2018-token} & $0.01$ & 29.19 & 20.94 & 40.19\\
        \midrule
        BERT+BLEU4 & $0.1$ & \bf 29.99 & \bf 22.82 & 39.84\\
        BERT+BLEU4 & $0.01$ & 29.51 & 22.30 & \bf 40.82\\
        \bottomrule
    \end{tabular}
    }
    
    \caption{BLEU4 evaluation scores on translation tasks from different label smoothing methods. We ran a bootstrap test \citep{koehn-2004-statistical} for estimating the significance of improvement over the strongest baseline and found that on all three datasets the improvement is statistically significant, $p<0.05$.}
    \label{tbl:res}
\end{table*}

The elements of $\NCal( \yvec )$ can be expected to have high semantic similarity with $\yvec$, which is desirable.
However, such sequences may \emph{not} meaningfully align with the original input $\xvec$
(e.g., may not be sufficiently close translations).
To account for this, we prune the elements from $\NCal( \yvec )$ based on the BLEU score.
Intuitively, this pruning retains sequences 
that are both semantically similar 
\emph{and}
have non-trivial token overlap with $\yvec$.

We use $\YCal_{\mathrm{ref}}$ as all output sequences in the training set.
In practice, one may however use any set of sequences that are valid for the domain in question.
We find $k = 100$ closest sequences in this space,
and smooth over $k' = 5$ pruned sequences with the highest BLEU score to $\yvec$.
In Table~\ref{tbl:examples} we show example augmentations. 
Notice both the diversity of augmentations, as well as relatedness to the original targets.

\begin{table*}[!t]
    \centering
    {
    \begin{tabular}{lcccccc@{}}
        \toprule
        \multicolumn{1}{c}{} & BLEU3 & BLEU4 & BLEU5 & METEOR & ROUGE & CIDER\\
        \toprule
        
        \cite{elbayad-etal-2018-token} & 27.9 & 20.94 & 15.93 & 24.92 & 50.98 & 211.49\\
        BERT+BLEU4 & 29.8 & 22.82 & 17.73 & 26.03 & 52.29 & 228.26\\
        \bottomrule
    \end{tabular}
    }
    
    \caption{Comparison of our model against the strongest baseline \cite{elbayad-etal-2018-token} as reported in Table~\ref{tbl:res} on EN-CS across multiple metrics. }
    \label{tbl:more_metrics}
    \vspace{-\baselineskip}
\end{table*}

\section{Experiments}\label{sec:experiments}
\noindent \textbf{Setup.} We use the Transformer model for our experiments, and follow the experimental setup and hyperparameters from \citet{Vaswani:2017}.
We experiment on three popular machine translation tasks: English-German (EN-DE), English-Czech (EN-CS) and English-French (EN-FR), using the WMT training datasets, and on the tensor2tensor framework~\cite{vaswani-etal-2018-tensor2tensor}.\footnote{Data available at \url{https://tensorflow.github.io/tensor2tensor/}.}
We evaluate on the Newstest 2015 for EN-DE and EN-CS, and WMT 2014 for EN-FR.

\noindent \textbf{Baselines.} We use the seq2seq model results by~\citet{Vaswani:2017} as a baseline. We compare our approach with the following alternate smoothing methods: i) smoothing is done over all possible tokens from the vocabulary at each next token prediction~\citep{Szegedy:2016}, ii) smoothing is conducted over random targets from within batch~\cite{guo-etal-2018-effective}, and iii) smoothing is done over artificially generated targets that are close to the actual target sequence according to BLEU score~\cite{elbayad-etal-2018-token}. For all these methods we experiment with values of $\alpha$ in $\{0.1, 0.001, 0.0001, 0.00001\}$, and report the best results in each case.
For the \cite{elbayad-etal-2018-token} baseline, we follow the reported best performing variant, randomly swapping tokens with others from the target sequence.

\vspace{3pt}
\noindent \textbf{Main results.} In Table~\ref{tbl:res} we report results from our method (BERT+BLEU) and the different state-of-the-art methods mentioned above. Our most direct comparison is against~\cite{elbayad-etal-2018-token}, as both the methods smooth over sequences that have high BLEU score. However, instead of generating sequences by randomly replacing tokens, we retrieve them from a corpus of well formed text sequences. In particular, we use BERT-base multilingual model to embed all the training target sequences into 768 dim fixed vector representation (corresponding to CLS token) and then identify top-100 nearest neighbors for each of the target sequence. Consequently, our method outperforms~\cite{elbayad-etal-2018-token} by a large margin on all three benchmarks. 
This demonstrates the importance of smoothing over sequences that not only have significant n-gram overlap with the ground truth target sequence but are also well formed and are semantically similar to the ground truth.
In Table~\ref{tbl:more_metrics} we report the comparison between our model and the strongest baseline on EN-CS across multiple metrics, confirming the improvement we report in Table~\ref{tbl:res} for BLEU score.

\begin{table}[!t]
    \centering
    {
    \begin{tabular}{lccc@{}}
        \toprule
        \multicolumn{1}{c}{} & BLEU3 & BLEU4 & BLEU5\\
        \toprule
        BERT+BLEU3 & 29.12 & 22.03 & 16.89\\
        BERT+BLEU4 & \bf 29.80 & \bf 22.82 & \bf 17.73\\
        BERT+BLEU5 & 29.41 & 22.38 & 17.26\\
        \bottomrule
    \end{tabular}
    }
    
    \caption{Results on EN-CS from targets smoothing with varying n-gram overlap enforced for the final selection of top $5$ augmented targets. Enforcing higher overlap to the original target worsens the performance.}
    \label{tbl:ablation}
    \vspace{-\baselineskip}
\end{table}

\noindent {\bf Ablating BLEU pruning.} Table~\ref{tbl:ablation} reveals it is useful to use a sufficiently restrictive criterion in BLEU pruning;
however, excess pruning (BLEU5) is harmful.
Thus, we seek to retrieve semantically related targets which do not necessarily have highest scoring n-gram overlap to the original target. 
This is intuitive:
enforcing too high n-gram overlap may cause all augmented targets to be too lexically similar, limiting the benefit of seeing new targets in training. 
We also experimented with not reranking neighbors using BLEU pruning, which resulted in no improvement over the baseline. In other words, it was essential to use this kind of postprocessing for obtaining improvements.

\noindent {\bf Ablating the number of neighbors.} 
We experimented with how the number of neighbors influences the results. For EN-CS, we obtained the following BLEU4 scores correspondingly for 10, 5 and 3 neighbors: 21.86, 22.82, 22.23. Overall, we find that too few or too many neighbors harm the performance compared to the 5 neighbors we used in other experiments. At the same time, the time complexity increases linearly as number of neighbors increases.

\section{Conclusion}
We propose a novel label smoothing approach for sequence to sequence problems that selects a subset of sequences 
that are not only semantically similar to the target sequences,
but are also \emph{well formed}. 
We achieve this by using a pre-trained model to find semantically similar sequences from the training corpus, and then we use BLEU score to rerank the closest targets. Our method shows a consistent and significant improvement over state-of-the-art techniques across different datasets.

In future work, we plan to apply our semantic label smoothing technique to various sequence to sequence problems, including Text Summarization \cite{zhang19} and Text Segmentation \cite{lukasik2020text}.
We also plan to study the relation between pretraining and data augmentation techniques.

\bibliography{main}
\bibliographystyle{acl_natbib}

\end{document}